\documentclass[conference]{IEEEtran}
\IEEEoverridecommandlockouts
\usepackage{cite}
\usepackage{amsmath,amssymb,amsfonts}
\usepackage{algorithmic}
\usepackage{graphicx}
\usepackage{textcomp}
\usepackage{xcolor}
\usepackage[utf8]{inputenc}
\usepackage{float}
\usepackage{balance}
\usepackage{url}

\def\BibTeX{{\rm B\kern-.05em{\sc i\kern-.025em b}\kern-.08em
    T\kern-.1667em\lower.7ex\hbox{E}\kern-.125emX}}

\usepackage{array,multirow}
\usepackage{booktabs}
\usepackage{pifont}

\title{Fidel: Reconstructing Private Training Samples from Weight Updates in Federated Learning}

\author{
\IEEEauthorblockN{David Enthoven \ \ \ \ \ \ \ \ Zaid Al-Ars}
\IEEEauthorblockA{\textit{Accelerated Big Data Systems Group} \\
\textit{Delft University of Technology, NL}\\
\{D.A.Enthoven, Z.Al-Ars\}@tudelft.nl}
}

\begin{document}
\maketitle

\begin{abstract}
With the increasing number of data collectors such as smartphones, immense amounts of data are available. Federated learning was developed to allow for distributed learning on a massive scale whilst still protecting each users’ privacy. This privacy is claimed by the notion that the centralized server does not have any access to a client’s data, solely the client’s model update. In this paper, we evaluate a novel attack method within regular federated learning which we name the \emph{First Dense Layer Attack (Fidel)}. The methodology of using this attack is discussed, and as a proof of viability we show how this attack method can be used to great effect for densely connected networks and convolutional neural networks. We evaluate some key design decisions and show that the usage of ReLu and Dropout are detrimental to the privacy of a client's local dataset. We show how to recover on average twenty out of thirty private data samples from a client’s model update employing a fully connected neural network with very little computational resources required. Similarly, we show that over thirteen out of twenty samples can be recovered from a convolutional neural network update. An open source implementation of this attack can be found here \url{https://github.com/Davidenthoven/Fidel-Reconstruction-Demo}
\end{abstract}


\section{Introduction} \label{sec:intro}
Deep learning algorithms have grown significantly in capabilities and popularity in the past decade. Advances in computing performance, increased data availability (and decreased data-storage costs) and developments of effective algorithms have created a surge of new applications using deep learning algorithms.

These algorithms commonly require an extensive amount of task-specific data to be accurate. Conventionally, this data is gathered and stored on an accessible centralized server, which is becoming increasingly cheap to access due to the increasing bandwidths available for remote data storage and retrieval.

This centralized data storage, however, introduces two problems. First, because of the exponential increase of data collectors in the field, the amount of available data is becoming excessively large, which increases the bandwidth requirements for communication as well as the computational requirements needed for processing. Second, devices may gather data of sensitive nature subject to privacy laws and regulations~\cite{GDPR}. Therefore, it is important to manage this sensitive data in a way that protects privacy. To combat these problems, alternative ways of decentralized learning have been proposed, most notably that of federated learning (FL)~\cite{FedLearning}.

\begin{figure}
    \centerline{\includegraphics[width=\linewidth]{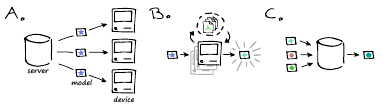}}
    \caption{The federated learning process: A) A model is sent from the server to the client devices. B) The clients train the model further on their local dataset. C) The clients send their local model back to the server which combines them into a single new model.}
    \label{fig:fedsteps}
\end{figure}

FL is a specific form of distributed learning in which no actual data is sent to a centralized server. Instead, an abstraction of this data in the form of machine learning model weights or weight gradients is sent to the server. These model weights are trained locally on client devices, each with their respective datasets. After training, each client sends its updated model back to the centralized server. The server then aggregates the models into a single model which is commonly done by averaging the models. Hereafter, the aggregated model is sent back to the clients. Figure~\ref{fig:fedsteps} illustrates these steps. This process is performed multiple times iteratively, whereby all clients jointly improve a shared model. FL claims to have distinct privacy advantages over traditional centralized data-driven model training approaches as well as reducing communication bandwidth requirements~\cite{FedLearning}. FL protects sensitive data from external adversaries, but it does not guarantee safety against internal adversaries such as a malicious centralized server. 

Federated learning provides substantial benefits compared to traditional machine learning in certain cases. Exemplary implementations for next word prediction on mobile devices~\cite{Fed_nextword, Fed_nextword2} and emoticon prediction~\cite{Fed_emoticon} show promising results for the use case of mobile phones. FL has found its use in medical applications as well~\cite{eurocat,stefan2020,medfed1,medfed2,medfed3,medfed4,medfed5} which creates a need for justification of the security of FL because of the sensitive nature of medical data. This security comes two-fold. First, FL should protect against adversarial clients. Second, modern information storage legislation such as the European General Data Protection Regulation (GDPR~\cite{GDPR}) stipulates that user data cannot be gathered without the users' consent. Therefore, FL must be demonstrably secure against the server obtaining private user information from the user models.

Recent research has shown multiple vulnerabilities in the privacy protection capabilities of FL~\cite{enthoven2020overview, threatsurvey1,DLG,iDLG,First-layer-attack,warfarin,MIA1,mGAN-AI,GAN-attack,backdoor2,Backdoorattack,backdoor3,sybilattack}. 
Identifying new attack methods and analyzing their capabilities and limitations is of high importance to understand their impact and to propose defensive strategies against them~\cite{enthoven2020overview}.

In this paper, we discuss a mechanism by which a deep fully connected neural network trained on a singular sample can be expected to reveal this underlying training sample in full. We call this mechanism the \emph{First Dense Layer Attack (Fidel)}. Then, we describe a methodology of how to apply the Fidel and subsequently, thoroughly demonstrate the efficacy of the Fidel for cases of both \emph{fully connected neural networks (FCNNs)} and \emph{convolutional neural networks (CNNs)}. This demonstration includes a practical study of applicability concerning various activation functions, model design decisions and local dataset sizes. We formulate key insights which serve as a starting point for extending the application domain of this attack method.

This paper is organized as follows. Section~\ref{sec:Fidel} describes the mathematical foundation of our method of attack and discusses the most important exploitable surface concerning federated learning. Section~\ref{sec:Fidel_methodology} discusses the methodology of how to apply Fidel in a practical setting. Hereafter a demonstration is given on trivial one-sample training rounds in both a FCNN and CNN. Section~\ref{sec:EXP} demonstrates the efficacy of this attack for more realistic multi-sample training rounds. Section~\ref{sec:gradients} provides insight into how this attack can be used to its fullest extent, as well as how protection against this attack can be achieved. Section~\ref{sec:discussion} provides a selection of considerations essential to understanding Fidel. These include, among others, limitation in applications and considerations with regards to privacy. Finally, Section~\ref{sec:conc} ends with the conclusions.

\section{First Dense Layer Attack}\label{sec:Fidel}
In this Section, the mathematical foundation of the Fidel attack method is explained. Hereafter, the security implications that this may have on federated learning are briefly discussed.

\subsection{Mathematical foundation}
The mathematical foundation of this attack method is formulated by Le Trieu et al.~\cite{First-layer-attack} and is repeated in this section as background information. Le Trieu et al. found that neurons in densely connected layers can reconstruct the activation of the previous layer by leveraging the fact that the weight and bias changes associated with this neuron are relative to the activation of the previous layer.

\begin{figure}
    \centerline{\includegraphics[width=0.5\linewidth]{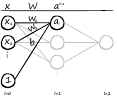}}
    \caption{Activation of a neuron in the first dense layer.}
    \label{fig:recnet}
\end{figure}

The activation of a neuron $i$ in a densely connected layer $l$ is formulated as $a_i = A(W^{l-1}_i \cdot a^{l-1}+b_i)$ in which $b_i$ is the bias of neuron $i$, $W^{l-1}_i$ is the weight vector connecting layer $l-1$ to neuron $i$ and $A()$ is the activation function. For a small network that connects an input vector to a singular neuron as illustrated in Figure~\ref{fig:recnet}, the activation of the previous layer is the actual input values $x$ of a sample thus $a_i = A(W\cdot x+b)$. Generally, for a given input $x$ we note that $a^0 = x$ and $h(x)$ is the activation of the last layer of the network.

Loss on a training sample $(x,y)$ is calculated by applying a loss function $\mathcal{L}$ on the network with its weights $W$ biases $b$. The loss determines the error of the weights based on the distance between the true value $y$ and the output value for $x$ as calculated by the network $h(x)$. For a single sample $(x,y)$ the loss function is $\mathcal{L}(W,b,x,y)$.

Two commonly used loss functions on which this attack works are squared error and cross-entropy loss functions. Using the squared error function as an example, the loss is calculated as $\mathcal{L}(W,b,x,y) = (h(x)-y)^2$.

To find the numerical direction in which the weights and biases of the network must move to provide a lower loss value for a specific training sample $(x,y)$, the gradient of each weight and bias are calculated. 
\begin{displaymath}g_{w,j}   = \frac{d\mathcal{L}(W,b,x,y)}{dW}=2(h(x)-y)A'(W\cdot x+b)\cdot x_j\end{displaymath}
\begin{displaymath}g_b       = \frac{d\mathcal{L}(W,b,x,y)}{db}=2(h(x)-y)A'(W\cdot x+b)\cdot 1\end{displaymath}
where $j$ is the $j^{th}$ input element in the input space of $x$.

\begin{figure}
    \centerline{\includegraphics[width=\linewidth]{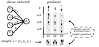}}
    \caption{Weight gradient on different samples (length of arrow represents value)}
    \label{fig:mathillu}
\end{figure}

We can identify two important characteristics of this derivation. First, having $g_w$ and $g_b$ allows us to retrieve the entire input $x$ on which the network was trained by dividing $\frac{g_w}{g_b}=x$. Secondly, if $g_b$ is not known, $g_w$ allows for a reconstruction of the input $x$ that is linearly related to the true sample $x$. This principle is illustrated in Figure~\ref{fig:mathillu} in which three different samples can be derived from their respective training gradients even if the bias gradient is unknown.

After the gradients of a model are calculated (can be aggregation over multiple samples) the weights of the model are updated by $w_{t+1} \leftarrow w_t - \eta g$ in which $\eta$ represents the so called learning rate. Having the before and after weights ${w_{t+1},w_t}$ or the difference in weight value $\delta = w_{t+1} - w_{t}$ can result in the same reconstruction for $x$ as described above since $$x = \frac{g_w}{g_b} = \frac{\frac{(w_t - w_{t+1})_w}{\eta}}{\frac{(w_t - w_{t+1})_b}{\eta}} = \frac{ (w_t - w_{t+1})_w}{ (w_t - w_{t+1})_b}=\frac{-\delta_w}{-\delta_b}$$

\subsection{Exploitable surface in federated learning}
The previous section described how a neuron on any dense layer can reconstruct the activation of the previous layer if either the gradients of this layer or any form of weight update is known.

This can be exploited during regular federated learning operation since the server must receive information about the change of weights of the participating clients. This exploitable surface makes it so that federated learning in its basic form is fundamentally unsafe against MITM attack and server-side attacks. Using more training-samples per update or a different network architecture such as CNN is no guarantee against this attack as will be demonstrated later on in Section~\ref{sec:EXP}.

Commonly, dense layers have more than a singular neuron, thus multiple reconstructions of the same input can be reconstructed. This induces reconstruction robustness against noise, lossy compression or other model update perturbation. Additionally, communication bandwidth reduction techniques that rely on communicating randomly chosen weights or other partial weight update omittance schemes do not guarantee safety against this attack. Multiple neurons can be employed in order to collaboratively reconstruct the original sample. 

\section{Fidel attack methodology}\label{sec:Fidel_methodology}

The Fidel allows a server or an adversary impersonating a server to gain access to data which it should not have. For this analysis, we adopt the \emph{honest-but-curious server} threat model~\cite{enthoven2020overview}
This attack method operates passively on the server-side, which means that it is hard to detect. As discussed in this section, the implementation is easy and can produce accurate reconstructions of a victim client's private training data.

\begin{figure}
    \centerline{\includegraphics[width=\linewidth]{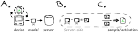}}
    \caption{Steps for Fidel method. A) the device trains on a private sample and sends its updated model to the server. B) the server subtracts the client's updated model from the client's previous model. C) The server divides the weight change of a neuron in the first dense layer by its bias change. }
    \label{fig:fidelsteps}
\end{figure}

In a federated setting, this attack is easy to execute. Consider a client that is trained on a single sample for one epoch. The client sends its updated model to the server. The server picks a neuron in the first dense layer (of the client's model) and divides the change in weights values weights by the respective change in bias value (which is part of the model update) revealing the training sample in full. This process is illustrated in Figure~\ref{fig:fidelsteps}.

The weights leading up to a neuron in a densely connected layer reveal the activation of the previous layer. This is the input sample for a FCNN but for a network that has convolutional layers before its first dense layer, the attack simply reveals the activation of this convolutional layer.

Importantly, such an input reconstruction may be performed for each neuron in the first dense layer. This has huge implications for the privacy of the underlying data. We call the reconstruction based on a single neuron \emph{partial reconstruction}. Networks with convolutional layers commonly use multiple convolution filters in parallel which means that for CNN's the partial reconstructions have channels equal to the number of filter kernels.

Included in the methodology of Fidel is the notion that the adversary can reconstruct the original sample by means of deconvoluting and up-scaling partial reconstructions. In this paper, we show the applicability of generative neural networks to do so. 

\subsection{FCNN single sample demonstration}
\begin{figure}
    \centerline{\includegraphics[width=\linewidth]{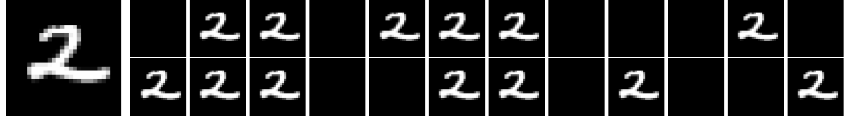}}
    \vspace{2px}
    \centerline{\includegraphics[width=\linewidth]{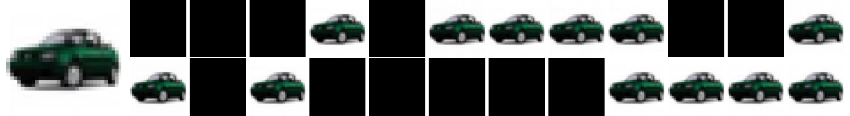}}
    \caption{Private sample and corresponding first 24 partial reconstructions. Top: MNIST, Bottom: CIFAR-10}
    \label{fig:DNN_single}
\end{figure}

In an FCNN, the first layer is densely connected to the input. In this experiment, the first layer consists of 128 neurons. Therefore, 128 partial reconstructions are available to the adversary. After training the network for some rounds on the training-set, the client trains on a singular training sample of the test-set. By executing the Fidel on the weight update of the model, the adversary now has 128 partial reconstructions available. The first 24 partial reconstructions for the MNIST and CIFAR-10 datasets are illustrated in Figure~\ref{fig:DNN_single} in combination with the private training sample. The illustrated partial reconstructions values are normalized to be displayed properly.

It is clearly visible how such reconstructions reveal the underlying data. Note that some partial reconstructions are fully black. This is because of the application of the \emph{ReLu activation function}; if the sum of weighted inputs is negative, the weights of this neuron will have no gradient and no reconstruction can be made from this neuron.

\subsection{CNN single sample demonstration}
\begin{figure}
    \centerline{\includegraphics[width=\linewidth]{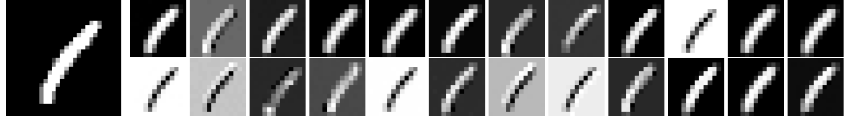}}
    \vspace{2px}
    \centerline{\includegraphics[width=\linewidth]{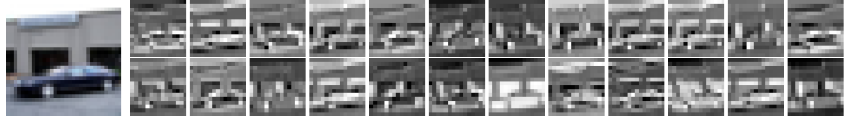}}
    \caption{Private sample and corresponding first 24 channels of 1 partial reconstruction a single neuron in the first dense layer. Top: MNIST, Bottom: CIFAR-10}
    \label{fig:CNN_single}
\end{figure}

Neural networks trained for image recognition often implement convolutional layers followed by max-pooling layers. The partial reconstructions for a CNN are thus the convoluted and subsequently max-pooled input samples. The partial reconstructions for the first 24 filters of the first neuron are illustrated in Figure~\ref{fig:CNN_single}. This small subset is only 24 of the possible 32 * 128 = 4096 (13x13 \& 15x15) partial reconstructions available to the adversary.

These partial reconstructions may reveal information outright. The MNIST reconstructions show this particularly well. This is because a convolution operation often retains some spatial information of their input. A larger amount of stacked convolutional layers may reduce the ability to reveal private information in partial reconstructions. Reconstructing the original sample from such partial reconstructions can be done in various ways. In this paper, we use a generative neural network as an example.

\begin{figure}
    \centerline{\includegraphics[width=\linewidth]{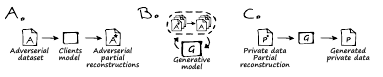}}
    \caption{Recreating training samples. A. Creating adversarial partial reconstructions. B. Training a generative network to reconstruct original samples from their partial reconstructions. C. Use the generative network to reconstruct the victim's samples.}
    \label{fig:cnngen}
\end{figure}

\paragraph*{Generative network}
In federated learning, the entire model including convolution filters is available to the server. Therefore, a generative neural network may be employed to piece together the original training sample from its convoluted and max-pooled partial reconstructions. A step-wise approach is taken which is illustrated in Figure~\ref{fig:cnngen}.

First, the adversary needs to have access to an auxiliary dataset. We can expect the best results when this dataset is stylistically similar to the training samples of the victim. This dataset can, for instance, be a test set available to the server. Using the client model, a set of corresponding partial reconstructions is created for each auxiliary dataset sample. The auxiliary dataset sample and its partial reconstructions are stored as a corresponding pair.
A generative neural network is then trained with the goal of recreating the original auxiliary dataset sample from the associated partial reconstructions. Hereafter, the generative network reconstructs the victim's private sample. The victim’s partial reconstructions (obtained with the Fidel) are used as an input on this generative neural network to reconstruct the victim's original training sample.

\begin{figure}
    \centerline{\includegraphics[width=\linewidth]{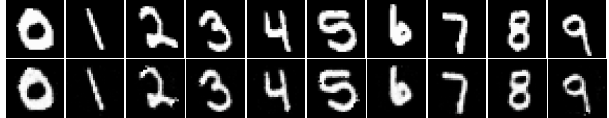}}
    \vspace{2px}
    \centerline{\includegraphics[width=\linewidth]{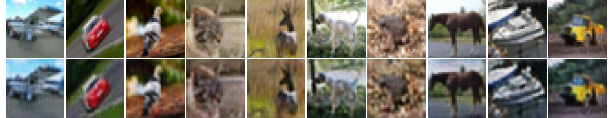}}
    \caption{Input samples (upper row) and their respective generated samples (lower row) with the generative network approach. Top 2 rows: MNIST, Bottom 2 rows: CIFAR-10}
    \label{fig:cnngen_sing}
\end{figure}

This approach is demonstrated with the use of a small (non-optimized) generative neural network described in Table~\ref{tab:mnist-gen} for the MNIST generative network and in Table~\ref{tab:cifar-gen} for the CIFAR-10 generative network. These networks were trained no longer than 10 minutes each. Figure~\ref{fig:cnngen_sing} shows 10 exemplary private samples of a client and their associated reconstructions. In this demonstration, the first 6000 images of the test set of MNIST and CIFAR-10 are used as the auxiliary dataset and the private samples are chosen from the other 4000 test set images.

It is directly visible that the reconstructions reveal close to all private information of the original sample. Noticeable differences between the ground truth and reconstruction are that for MNIST some regions seem noisy and pixelated and for CIFAR-10 the reconstructions are blurrier.

We justify the notion that this technique is also viable on more complex networks by pointing to our methods resemblance to an auto-encoder network. In such a network, the victims model intuitively serves as the (static) encoder part and the generative adversarial network serves as a trainable decoding structure.

\section{Experimental Results}\label{sec:EXP}
This section describes the experimental results and findings by performing the Fidel on simulated clients that have larger local datasets.

\subsection{Experimental setup}
For an FCNN, the gradient update of each neuron in the first layer can provide a full reconstruction of the provided input sample. Figure~\ref{fig:recnet} illustrates the first layer of such a network. Commonly such a network has more than one neuron indicating that multiple reconstructions are possible for each weight update.

We use two basic networks for the experiments in this paper as the target victim networks, represented by an FCNN as described in Table~\ref{tab:ArchFCNN}, and a CNN as described in Table~\ref{tab:ArchCNN}. These networks are used for demonstrative purposes and are not optimized for any specific application.

We demonstrate the experiments on the MNIST~\cite{MNIST_Data} and CIFAR-10~\cite{Cifar10_Data} datasets which consist of 70k 28x28 gray-scale and 60k 32x32 RGB images, respectively. Among these images, we reserve 10k images as a test set in both datasets. These datasets are intended for image classification tasks with 10 distinct labels.

\subsection{FCNN multi-sample reconstruction}
\begin{figure}
    \centerline{\includegraphics[width=\linewidth]{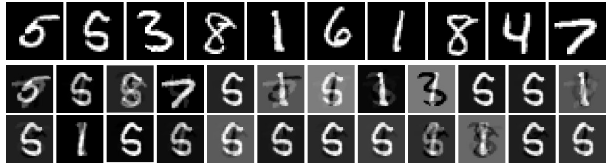}}
    \vspace{2px}
    \centerline{\includegraphics[width=\linewidth]{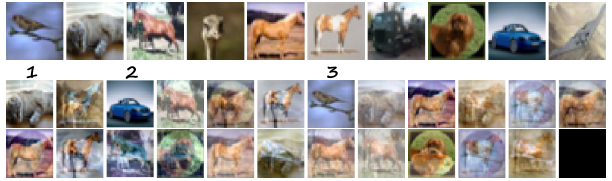}}
    \caption{Batch of 10 samples and 24 partial reconstructions. Top 3 rows: MNIST, Bottom 3 rows: CIFAR-10}
    \label{fig:FCNN_batch10}
\end{figure}

In this section, we illustrate the first 24 partial reconstructions out of a possible 128 formed from the weight update of a client's network trained for one epoch on a batch of 10 samples. Applying the attack yields partial reconstructions as illustrated in Figure~\ref{fig:FCNN_batch10}

From these examples, it is possible to see that even for models trained on more than only one sample, information is still revealed. For the CIFAR-10 example, the 3 partial reconstructions annotated by numbers (1, 2 and 3) are particularly revealing of the original private samples. It is interesting to note that in the figure, there is one neuron that did not activate for any of the provided samples and thus gave a fully black reconstruction.

Without any additive noise or other data modification, the partial reconstructions are combinations of the input images individually scaled with some (possibly negative) factor. Therefore, theoretically, these partial reconstructions can be used to retrieve all underlying samples employing blind source separation algorithms. We do not explore this notion further in this paper.

\subsection{CNN multi-sample reconstruction}
\begin{figure}
    \centerline{\includegraphics[width=\linewidth]{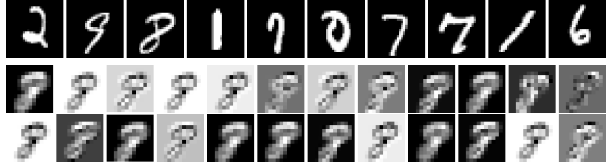}}
    \vspace{2px}
    \centerline{\includegraphics[width=\linewidth]{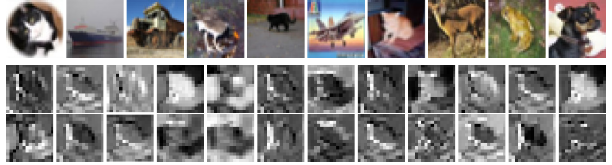}}
    \caption{Input batch of 10 training samples and its respective set of 24 partial reconstructions from one neuron. Top 3 rows: MNIST, Bottom 3 rows: CIFAR-10}
    \label{fig:CNN_batch10}
\end{figure}

Figure~\ref{fig:CNN_batch10} shows the first 24 out of 32 filters for a single neuron on a model training update of 10 samples. Note that the neuron is particularly responsive to the 3rd sample for MNIST and the 7th sample for CIFAR-10. 

Using the same generative method as demonstrated above, we show generated reconstructions from a local private data set of 10 samples. Each set of partial reconstructions retrieved associated with a neuron is now used to generate possible reconstructions. In Figure~\ref{fig:cnngen_batch}, 100 reconstructions for an MNIST sample batch and 120 reconstructions for a CIFAR-10 sample batch are illustrated. Within these reconstructions, nearly all private samples can be identified (albeit with varying degrees of fidelity). 

\section{Gradients in Fidel}\label{sec:gradients}
In this section, the underlying principles of this attack method are explained.

\subsection{Negligible gradients}
\begin{figure}
    \centerline{\includegraphics[width=0.8\linewidth]{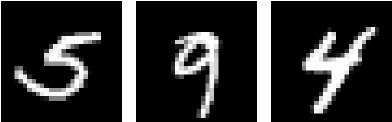}}
    \vspace{5px}
    \centerline{\includegraphics[width=0.8\linewidth]{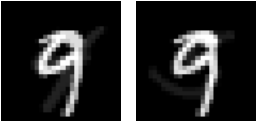}}
    \caption{Top: Private training data, Bottom: 2 partial reconstructions. 
    In the partial reconstructions the original samples 4 and 5 are faintly visible whereas the 9 is completely recognizable.}
    \label{fig:neg_gradients}
\end{figure}

To understand why it is possible to see fully reproduce images when training on a batch of multiple images, we introduce the term \emph{negligible gradients}. This term is used to identify a gradient of a sample (for a single neuron) which is an order of magnitude smaller than another sample gradient update. A demonstration of this is given in Figure~\ref{fig:neg_gradients}. In this figure, a client trained on the FCNN network with 3 local samples (at the top) can leak (among others) the shown two partial reconstructions (at the bottom). Within these partial reconstructions, the number four and five are visible but do not obfuscate any private information about the number nine.
 
In Figure~\ref{fig:FCNN_batch10}, we showed that for a batch of 10 samples it is relatively likely that in an FCNN a neuron is only activated on a single sample of the whole batch. This phenomenon propagates with the CNN network where a neuron in the first dense layer (after the convolution layers) only activates for a single sample, which in turn allows for reconstruction as shown in Figure~\ref{fig:CNN_batch10}.

Two noteworthy factors influence this phenomenon to the detriment of privacy in the federated learning process. First, the usage of the ReLu activation function greatly enhances the likelihood that a neuron is only activated on a single sample. The absence of positive value ceilings in ReLu increases the likelihood that the activation of a sample is so large that all other samples have negligible gradient updates on its weights. Second, the usage of dropout after the first dense layer also increases the likelihood that a neuron is only activated for a single sample in the batch. Because dropout forces a samples gradient to zero for randomly chosen neurons.

\subsection{Fully revealed private samples}
For image data, evaluating if private information is leaked is often not dependent on the value of single pixels. The context of the rest of the image can often provide enough information to compensate for the erroneous pixels. The reconstruction obtained with the Fidel may be scaled by some unknown (possibly negative) value.

To evaluate if a reconstruction is fully revealing a private sample, we use the Pearson correlation coefficients~\cite{dekking2006modern} between the reconstruction and the original sample. This metric is invariant against scaling of all sample values by some constant as well as invariant against adding some constant to all values in the sample. This metric is used to demonstrate that a linear relationship exists between the private sample and the reconstructed sample.

As a threshold for the complete reveal of a sample, we chose a correlation coefficient of 0.98 to indicate that all private information of the input image is fully revealed. This threshold provides some robustness against negligible gradients and floating-point calculation errors. In a practical setting, two images taken right after one another with the same equipment can be expected to have at best such a correlation coefficient~\cite{crosscorr_ineffective}.

\begin{figure}
    \centerline{\includegraphics[width=0.8\linewidth]{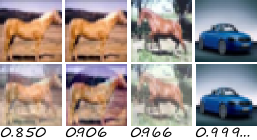}}
    \caption{Top: ground truth, Middle: reconstruction, bottom correlation coefficient between the two (samples taken from the experiment displayed in Figure~\ref{fig:FCNN_batch10}}
    \label{fig:corrcoeff_example}
\end{figure}

To justify this threshold Figure~\ref{fig:corrcoeff_example} shows example correlation coefficients between private samples and reconstructions in the batch FCNN setting. As depicted in this figure, correlations lower than 0.98 may still reveal sensitive private information, especially for image data.

\begin{figure}
    \centerline{\includegraphics[width=\linewidth]{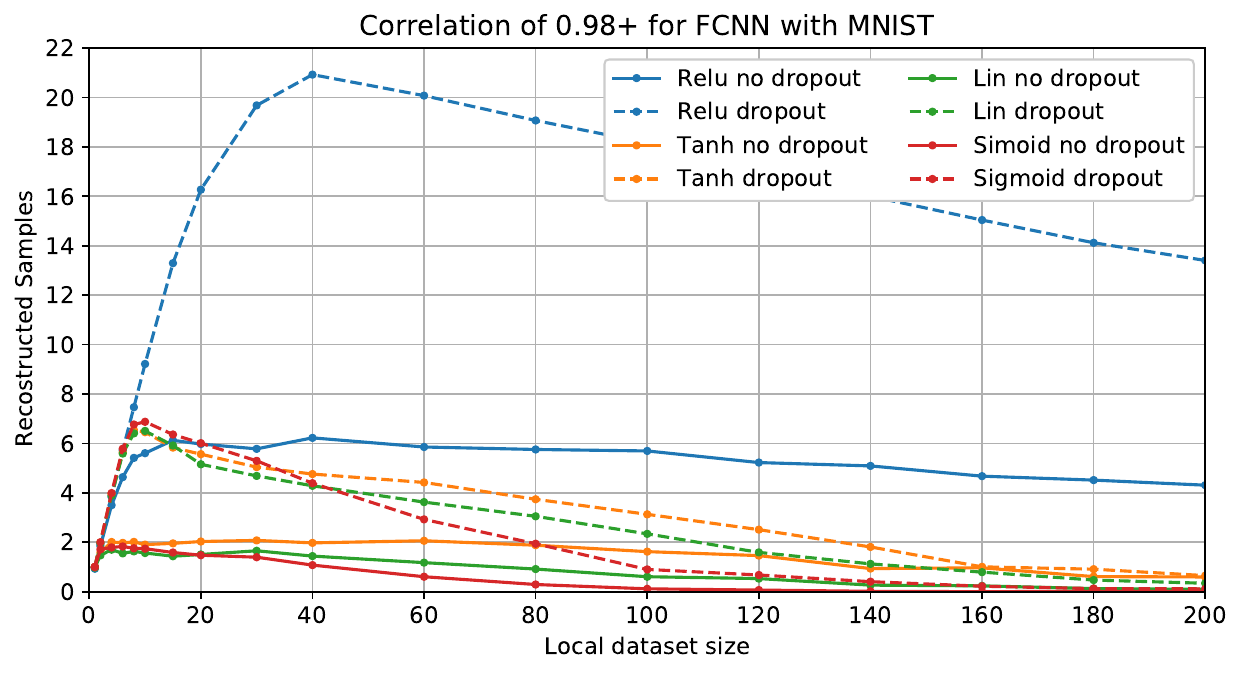}}
    \caption{Fully revealed private samples for the FCNN model on the MNIST dataset. The graph depicts a 200 measurement average.}
    \label{fig:regcognise_FCNN}
\end{figure}

For the FCNN MNIST demonstration, we test varying numbers of private samples used to train the local model to see how many of the underlying samples are fully revealed in the weight update. These results are illustrated in Figure~\ref{fig:regcognise_FCNN}. Private samples are not counted multiple times for different neurons, so it may very well be that multiple neurons fully reveal the same underlying private sample. The graph describes the average revealed samples per round over 200 measurements. Indicating the average per round reveal over 200 rounds of federated training. Notably, the graph shows the average sample reveal that is greater than one for nearly all scenarios. This means that having a large local dataset does not ensure privacy preservation as data samples can still be revealed with local datasets as large as 200 samples. 

The amount of fully revealed private samples is incredibly depend on the employed activation function as well as the usage of dropout. The graph shows that in using the ReLu activation function, the amount of revealed private samples is significantly higher than other activation functions. Using dropout is demonstrably effective in allowing the Fidel attack to succeed. The worst-case scenario in this experiment shows that as much as $\frac{2}{3}$ of a local dataset of 30 samples can be reconstructed.

If the first dense layer has more neurons than in the network employed in this experiment, we expect that even more private samples may be revealed per weight update. This is based on the notion that the neurons may be expected to reveal samples in a non-dependent probabilistic fashion. Additionally, the probabilistic nature of a neuron activating for a singular training sample means that the expected amount of revealed samples will increase cumulatively over the number of training rounds.

\begin{figure}
    \centerline{\includegraphics[width=\linewidth]{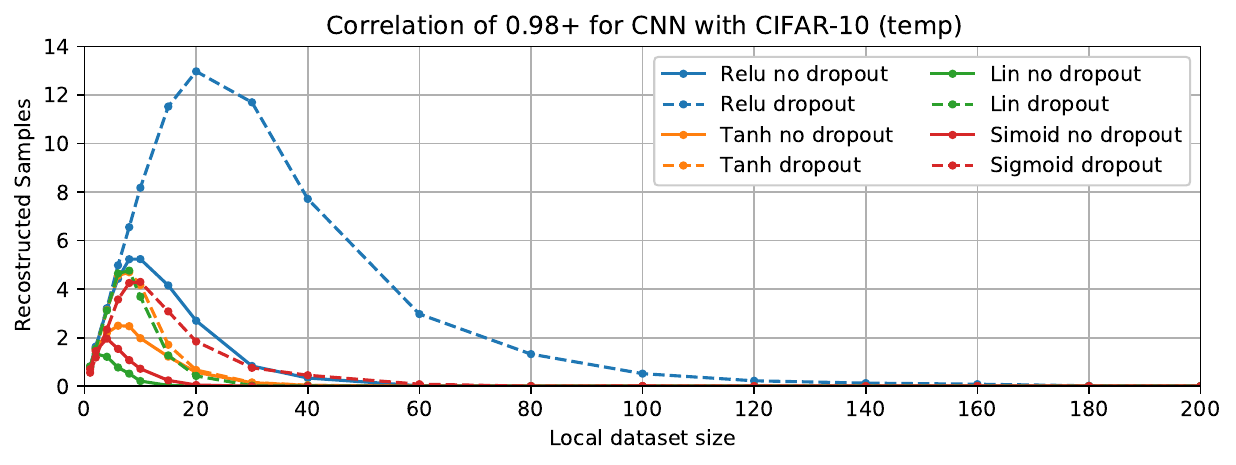}}
    \caption{Fully revealed private samples for the CNN model on the CIFAR-10 dataset employing a generative network for reconstruction. The graph depicts a 200 measurement average.}
    \label{fig:regcognise_CNN}
\end{figure}

We perform the same experiment on the generative method with the CNN network and the CIFAR-10 dataset. The results of this experiment are illustrated in Figure~\ref{fig:regcognise_CNN}. Here too, we find that especially for smaller sized local datasets the weight update may reveal private information that can be reconstructed to near-perfect accuracy. 

This graph shows that for the CNN network the local dataset size is of greater influence and the amount of fully revealed samples overall are far less than with the FCNN experiment. We can attribute this, in part, to the generative neural network that was employed for reconstruction. Additionally, the max-pooling layer might remove essential information needed for a full reconstruction.

From these measurements, it seems that under most circumstances the activation function and network design are of great influence to the preservation of privacy. It is important to re-emphasize that the graphs solely show fully revealed images for the chosen threshold of 0.98 correlation or more. As shown in Figure~\ref{fig:corrcoeff_example}, lower correlations may still reveal private information.

Federated learning has been designed specifically for the setting of numerous clients with small datasets. The threat of this attack may, therefore, be assessed among other things on the distribution of the data among clients i.e. the number of samples in a clients local dataset.

\section{Discussion} \label{sec:discussion}
The presented results are only accurate for the network in this experiment. These results are indicative of the exploitable potential in deep neural networks. This work of research is only meant to provide understanding about the methodology of this attack method and therefore serves as a starting point for more elaborate network-targeting attacks. For instance, the usage of blind source separation algorithms might provide a direct increase in the number of samples that can be reconstructed.

The Fidel is passively executed on the network, meaning that the training procedures are not influenced during the attack which makes it near undetectable. The application domain for this method of attack is for networks using the mean squared and categorical cross-entropy loss functions that use at least one densely connected layer. The same ideas may be used to broaden the application domain. One of the key strengths of this attack method is that the adversarial resource requirements are extremely low. For the FCNN setting a single divide, operation suffices in most cases. For the CNN setting the design of a generative network is required which may be subject to network-specific optimization.

The Fidel is a viable means of attack for most network designs. In networks employing ReLu Activation and Dropout layers, this attack is most effective. Furthermore, in settings where clients have small local data-sets, the Fidel proves additionally effective. This shows that this method of attack is particularly use-full for federated learning since one of federated learning's key implementation environments is where numerous clients each have a fraction of the data.

\paragraph*{Privacy}
One attractive feature of federated learning is that by design the server does not have access to the data but merely to an abstract representation of it. For companies, this means that there is little to no need of collecting informed consent from participating clients when the data is of personal nature. Recent legislation such as GDPR~\cite{GDPR} imposes strict rules on the collection and usage of personal data. The Fidel places a demonstrable burden on the deployment of federated learning by arguing that some private information can easily be recovered on the centralized server, which necessitates collecting the aforementioned client consent~\cite[Recital 26]{GDPR}. 

\section{Conclusion} \label{sec:conc}
For the use in federated learning, Fidel proves to be an extremely low-cost attack method that in cases may completely reveal all private training data from a participating client. We demonstrate this attack to be viable in FCNNs and CNNs. Reconstruction for a CNN network requires an additional (tailor-made) generative network as well as an auxiliary dataset. This work of research is for the purpose of demonstrating the Fidel methodology rather than setting a benchmark for reconstruction in federated learning.

Certain implementation considerations may greatly increase the viability of the Fidel. The most prominent is the usage of ReLu and Dropout layers. In addition, the Fidel benefits from small local client datasets, which is one of the key application purposes of federated learning. As such the Fidel can in cases nullify any privacy benefit gained from employing the federated learning strategy.

We demonstrate that on average twenty out of thirty private data samples from a client’s FCNN model update can be reconstructed. Additionally, over thirteen out of twenty samples can be recovered from a CNN update.

\section*{Acknowledgments}
This research was performed with the support of the European Commission, the ECSEL JU project NewControl (grant no. 826653), the Eureka PENTA project Vivaldy (grant no. PENT191004), and the Huawei grant no. YBN2019045113.

\bibliographystyle{unsrt}
\bibliography{References}

\begin{thebibliography}{10}

\bibitem{GDPR}
{Council of European Union}.
\newblock General data protection regulation, 2018.

\bibitem{FedLearning}
H.~Brendan McMahan, Eider Moore, Daniel Ramage, and Blaise~Ag{\"{u}}era
  y~Arcas.
\newblock Federated learning of deep networks using model averaging.
\newblock {\em CoRR}, abs/1602.05629, 2016.

\bibitem{Fed_nextword}
Andrew Hard, Kanishka Rao, Rajiv Mathews, Fran{\c{c}}oise Beaufays, Sean
  Augenstein, Hubert Eichner, Chlo{\'{e}} Kiddon, and Daniel Ramage.
\newblock Federated learning for mobile keyboard prediction.
\newblock {\em CoRR}, abs/1811.03604, 2018.

\bibitem{Fed_nextword2}
Timothy Yang, Galen Andrew, Hubert Eichner, Haicheng Sun, Wei Li, Nicholas
  Kong, Daniel Ramage, and Fran{\c{c}}oise Beaufays.
\newblock Applied federated learning: Improving google keyboard query
  suggestions.
\newblock {\em CoRR}, abs/1812.02903, 2018.

\bibitem{Fed_emoticon}
Swaroop Ramaswamy, Rajiv Mathews, Kanishka Rao, and Fran{\c{c}}oise Beaufays.
\newblock Federated learning for emoji prediction in a mobile keyboard.
\newblock {\em CoRR}, abs/1906.04329, 2019.

\bibitem{eurocat}
Timo~M. Deist, A.~Jochems, Johan van Soest, Georgi Nalbantov, Cary Oberije,
  Seán Walsh, Michael Eble, Paul Bulens, Philippe Coucke, Wim Dries, Andre
  Dekker, and Philippe Lambin.
\newblock Infrastructure and distributed learning methodology for
  privacy-preserving multi-centric rapid learning health care: eurocat.
\newblock {\em Clinical and Translational Radiation Oncology}, 4:24 -- 31,
  2017.

\bibitem{stefan2020}
Stefan Zwaard, Henk-Jan Boele, Hani Alers, Christos Strydis, Casey
  Lew-Williams, and Zaid Al-Ars.
\newblock Privacy-preserving object detection \& localization using distributed
  machine learning: A case study of infant eyeblink conditioning.
\newblock {\em arXiv:2010.07259}, 2020.

\bibitem{medfed1}
Arthur Jochems, Timo~M. Deist, Issam~El Naqa, Marc Kessler, Chuck Mayo, Jackson
  Reeves, Shruti Jolly, Martha Matuszak, Randall~Ten Haken, Johan van Soest,
  Cary Oberije, Corinne Faivre-Finn, Gareth Price, Dirk de~Ruysscher, Philippe
  Lambin, and Andre Dekker.
\newblock Developing and validating a survival prediction model for nsclc
  patients through distributed learning across 3 countries.
\newblock {\em International Journal of Radiation Oncology Biology Physics},
  99(2):344 -- 352, 2017.

\bibitem{medfed2}
Arthur Jochems, Timo~M. Deist, Johan van Soest, Michael Eble, Paul Bulens,
  Philippe Coucke, Wim Dries, Philippe Lambin, and Andre Dekker.
\newblock Distributed learning: Developing a predictive model based on data
  from multiple hospitals without data leaving the hospital – a real life
  proof of concept.
\newblock {\em Radiotherapy and Oncology}, 121(3):459 -- 467, 2016.

\bibitem{medfed3}
Li~Huang, Andrew~L. Shea, Huining Qian, Aditya Masurkar, Hao Deng, and Dianbo
  Liu.
\newblock Patient clustering improves efficiency of federated machine learning
  to predict mortality and hospital stay time using distributed electronic
  medical records.
\newblock {\em Journal of Biomedical Informatics}, 99:103291, 2019.

\bibitem{medfed4}
Theodora~S. Brisimi, Ruidi Chen, Theofanie Mela, Alex Olshevsky, Ioannis~Ch.
  Paschalidis, and Wei Shi.
\newblock Federated learning of predictive models from federated electronic
  health records.
\newblock {\em International Journal of Medical Informatics}, 112:59 -- 67,
  2018.

\bibitem{medfed5}
Micah~J. Sheller, G.~Anthony Reina, Brandon Edwards, Jason Martin, and Spyridon
  Bakas.
\newblock Multi-institutional deep learning modeling without sharing patient
  data: A feasibility study on brain tumor segmentation.
\newblock {\em Brainlesion : glioma, multiple sclerosis, stroke and traumatic
  brain injuries. BrainLes (Workshop)}, 11383:92--104, 2019.

\bibitem{enthoven2020overview}
David Enthoven and Zaid Al-Ars.
\newblock An overview of federated deep learning privacy attacks and defensive
  strategies.
\newblock {\em arXiv:2004.04676}, 2020.

\bibitem{threatsurvey1}
Lingjuan Lyu, Han Yu, and Qiang Yang.
\newblock Threats to federated learning: A survey, 2020.

\bibitem{DLG}
Ligeng Zhu, Zhijian Liu, and Song Han.
\newblock Deep leakage from gradients, 2019.

\bibitem{iDLG}
Bo~Zhao, Konda~Reddy Mopuri, and Hakan Bilen.
\newblock idlg: Improved deep leakage from gradients, 2020.

\bibitem{First-layer-attack}
Le~Trieu Phong, Yoshinori Aono, Takuya Hayashi, Lihua Wang, and Shiho Moriai.
\newblock Privacy-preserving deep learning: Revisited and enhanced.
\newblock In Lynn Batten, Dong~Seong Kim, Xuyun Zhang, and Gang Li, editors,
  {\em Applications and Techniques in Information Security}, pages 100--110.
  Springer Singapore, 2017.

\bibitem{warfarin}
Matthew Fredrikson, Eric Lantz, Somesh Jha, Simon Lin, David Page, and Thomas
  Ristenpart.
\newblock Privacy in pharmacogenetics: An end-to-end case study of personalized
  warfarin dosing.
\newblock In {\em Proceedings of the 23rd USENIX Conference on Security
  Symposium}, SEC'14, pages 17--32, Berkeley, CA, USA, 2014. USENIX
  Association.

\bibitem{MIA1}
Matt Fredrikson, Somesh Jha, and Thomas Ristenpart.
\newblock Model inversion attacks that exploit confidence information and basic
  countermeasures.
\newblock In {\em Proceedings of the 22Nd ACM SIGSAC Conference on Computer and
  Communications Security}, CCS '15, pages 1322--1333. ACM, 2015.

\bibitem{mGAN-AI}
Zhibo Wang, Mengkai Song, Zhifei Zhang, Yang Song, Qian Wang, and Hairong Qi.
\newblock Beyond inferring class representatives: User-level privacy leakage
  from federated learning.
\newblock {\em CoRR}, abs/1812.00535, 2018.

\bibitem{GAN-attack}
Briland Hitaj, Giuseppe Ateniese, and Fernando P{\'{e}}rez{-}Cruz.
\newblock Deep models under the {GAN:} information leakage from collaborative
  deep learning.
\newblock {\em CoRR}, abs/1702.07464, 2017.

\bibitem{backdoor2}
Xinyun Chen, Chang Liu, Bo~Li, Kimberly Lu, and Dawn Song.
\newblock Targeted backdoor attacks on deep learning systems using data
  poisoning.
\newblock {\em CoRR}, abs/1712.05526, 2017.

\bibitem{Backdoorattack}
Eugene Bagdasaryan, Andreas Veit, Yiqing Hua, Deborah Estrin, and Vitaly
  Shmatikov.
\newblock How to backdoor federated learning.
\newblock {\em CoRR}, abs/1807.00459, 2018.

\bibitem{backdoor3}
Tianyu Gu, Brendan Dolan{-}Gavitt, and Siddharth Garg.
\newblock Badnets: Identifying vulnerabilities in the machine learning model
  supply chain.
\newblock {\em CoRR}, abs/1708.06733, 2017.

\bibitem{sybilattack}
Clement Fung, Chris J.~M. Yoon, and Ivan Beschastnikh.
\newblock Mitigating sybils in federated learning poisoning.
\newblock {\em CoRR}, abs/1808.04866, 2018.

\bibitem{MNIST_Data}
Y.~{Lecun}, L.~{Bottou}, Y.~{Bengio}, and P.~{Haffner}.
\newblock Gradient-based learning applied to document recognition.
\newblock {\em Proceedings of the IEEE}, 86(11):2278--2324, Nov 1998.

\bibitem{Cifar10_Data}
Alex Krizhevsky.
\newblock Learning multiple layers of features from tiny images.
\newblock {\em University of Toronto}, 05 2012.

\bibitem{dekking2006modern}
F.M. Dekking, C.~Kraaikamp, H.P. Lopuha{\"a}, and L.E. Meester.
\newblock {\em A Modern Introduction to Probability and Statistics:
  Understanding Why and How}.
\newblock Springer Texts in Statistics. Springer London, 2006.

\bibitem{crosscorr_ineffective}
K.~Yen, Eugene~K. Yen, Roger~G. Johnston, Roger~G. Johnston, and Ph. D.
\newblock The ineffectiveness of the correlation coefficient for image
  comparisons.

\end{thebibliography}

\appendix
\section{Network details}\label{sec:appendix}
\begin{table}[H]
\caption{The architecture of FCNN network used for metric testing. (*) This activation function is changed in some experiments.}
\label{tab:ArchFCNN}
\begin{tabular}{l c c}
\toprule
layer           & size      & activation function \\ 
\midrule
(input)         & -          & -\\
Dense           & 128        & ReLu*\\
Dense           & 128        & ReLu\\
Dense           & 64         & ReLu\\
Dense           & 10         & Softmax\\
\midrule
\multicolumn{3}{l}{learning-rate:  0.01 }\\
\multicolumn{3}{l}{Loss function: categorical crossentropy }\\
\multicolumn{3}{l}{Optimizer: SGD with batch size 50}\\
\end{tabular}
\end{table}

\begin{table}[H]
\caption{The architecture of CNN network used for metric testing. (*) This activation function is changed in some experiments.}
\label{tab:ArchCNN}
\begin{tabular}{l c c}
\toprule
layer           & size      & activation function \\ 
\midrule
(input)         & -          & -\\
Conv 2D         & 3,3 X 32   & -\\
Maxpool         & 2,2        & -\\
flatten         & -          & - \\
Dense           & 128        & ReLu*\\
Dense           & 64          & ReLu\\
Dense           & 10         & Softmax\\
\midrule
\multicolumn{3}{l}{learning-rate:  0.01 }\\
\multicolumn{3}{l}{Loss function: categorical crossentropy }\\
\multicolumn{3}{l}{Optimizer: SGD with batch size 50}\\
\end{tabular}
\end{table}

\begin{table}[H]
\caption{The architecture of the generative neural network to recreate the MNIST samples from their original partial reconstructions.}
\label{tab:mnist-gen}
\begin{tabular}{l c l}
\toprule
layer                   &size              &activation function\\ 
\midrule
(input)                 &(13,13,32)              &-\\
up-sampling (nearest)   &2,2                     &-\\
Conv 2D                 &5,5 X 20                &ReLu\\
up-sampling (nearest)   &2,2                     &-\\
Conv 2D                 &5,5 X 10 stride(2,2)   &ReLu\\
flatten                 &-                      &-\\
dense                   &784                    &Sigmoid\\
reshape (28x28)         &(28,28)                &-\\
\midrule
\multicolumn{3}{l}{learning-rate: 0.001 rho: 0.95 }\\
\multicolumn{3}{l}{Loss function: mean squared error }\\
\multicolumn{3}{l}{Optimizer: Adadelta }\\
\end{tabular}
\end{table}

\begin{table}[H]
\caption{The architecture of the generative neural network to recreate the CIFAR-10 samples from their original partial reconstructions}
\label{tab:cifar-gen}
\begin{tabular}{l c c l l}
\toprule
layer           &size       &stride   &padding   &activation \\ 
\midrule
(input)         &(15,15,32) &-      &-      &- \\
ConvT 2D        &5,5 X 128  &1,1    &same   &Relu\\
ConvT 2D        &5,5 X 64   &2,2    &same   &Relu\\
Batchnorm       &-          &-      &-      &- \\
ConvT 2D        &5,5 X 64   &1,1    &same   &Relu\\
Conv 2D         &3,3 X 32   &1,1    &valid  &Tanh\\
Conv 2D         &3,3 X 3    &1,1    &same   &Sigmoid\\
\midrule
\multicolumn{5}{l}{learning-rate: 0.001 rho: 0.95}\\
\multicolumn{5}{l}{Loss function: mean squared error }\\
\multicolumn{5}{l}{Optimizer: adadelta}\\
\end{tabular}
\end{table}

\section{Reconstruction set}\label{sec:appendix2}

\begin{figure}[H]
    \centerline{\includegraphics[width=0.9\linewidth]{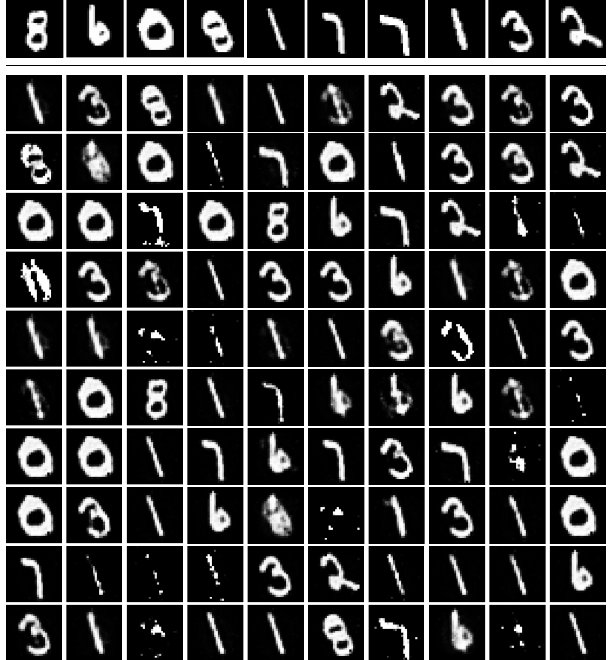}}
    \centerline{\includegraphics[width=0.9\linewidth]{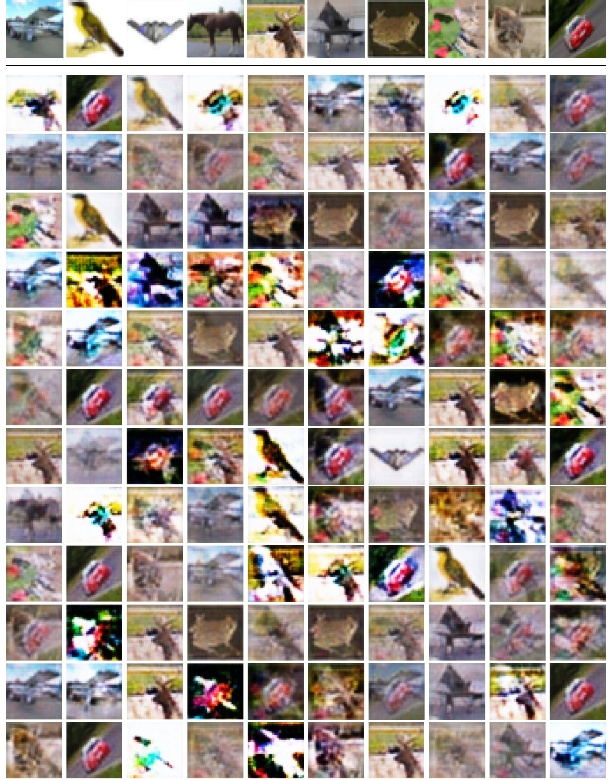}}
    \caption{Top. Batch of 10 private MNIST samples and 100 associated reconstructions.
    Bottom. Batch of 10 private CIFAR-10 samples and 120 associated reconstructions.
    These reconstructions are made by means of a generative neural network using the CNN partial reconstructions as an input.}
    \label{fig:cnngen_batch}
\end{figure}

\end{document}